\documentclass[10pt,twocolumn,letterpaper]{article}

\usepackage[pagenumbers]{cvpr} 

\usepackage{graphicx}
\usepackage{amsmath}
\usepackage{amssymb}
\usepackage{booktabs}
\usepackage{algorithm,algorithmic}
\usepackage{wrapfig}
\usepackage{caption}
\usepackage{subcaption}
\usepackage{sidecap}
\usepackage{multirow}
\usepackage{cuted}
\usepackage{makecell}

\usepackage[table]{xcolor} 
\usepackage{colortbl}      

\definecolor{mylightgreen}{HTML}{E8F5E9}

\newcommand{\modelname}{\textsc{R-C$^2$}\xspace}

\definecolor{cvprblue}{rgb}{0.21,0.49,0.74}
\usepackage[pagebackref,breaklinks,colorlinks,allcolors=cvprblue]{hyperref}

\def\paperID{8114} 
\def\confName{CVPR}
\def\confYear{2026}

\title{\modelname: Cycle-Consistent Reinforcement Learning Improves Multimodal Reasoning}

\author{
\textbf{Zirui Zhang$^{1}$}\quad
\textbf{Haoyu Dong$^{2}$}\quad
\textbf{Kexin Pei$^{3}$}\quad
\textbf{Chengzhi Mao$^{1}$} \\
$^{1}$Rutgers University\quad
$^{2}$Columbia University\quad
$^{3}$University of Chicago \\
{\tt\small https://zirui00.github.io/RC2-Project-Page/}
}

\begin{document}
\maketitle
\def\Blue{\color{blue}}
\def\Purple{\color{purple}}

\def\A{{\bf A}}
\def\a{{\bf a}}
\def\B{{\bf B}}
\def\b{{\bf b}}
\def\C{{\bf C}}
\def\c{{\bf c}}
\def\D{{\bf D}}
\def\d{{\bf d}}
\def\E{{\bf E}}
\def\e{{\bf e}}
\def\f{{\bf f}}
\def\F{{\bf F}}
\def\K{{\bf K}}
\def\k{{\bf k}}
\def\L{{\bf L}}
\def\H{{\bf H}}
\def\h{{\bf h}}
\def\G{{\bf G}}
\def\g{{\bf g}}
\def\I{{\bf I}}
\def\R{{\bf R}}
\def\X{{\bf X}}
\def\Y{{\bf Y}}
\def\OO{{\bf O}}
\def\oo{{\bf o}}
\def\P{{\bf P}}
\def\Q{{\bf Q}}
\def\q{{\bf q}}
\def\r{{\bf r}}
\def\s{{\bf s}}
\def\S{{\bf S}}
\def\t{{\bf t}}
\def\T{{\bf T}}
\def\x{{\bf x}}
\def\y{{\bf y}}
\def\z{{\bf z}}
\def\Z{{\bf Z}}
\def\M{{\bf M}}
\def\m{{\bf m}}
\def\n{{\bf n}}
\def\U{{\bf U}}
\def\u{{\bf u}}
\def\V{{\bf V}}
\def\v{{\bf v}}
\def\W{{\bf W}}
\def\w{{\bf w}}
\def\0{{\bf 0}}
\def\1{{\bf 1}}
\def\N{{\bf N}}

\def\AM{{\mathcal A}}
\def\EM{{\mathcal E}}
\def\FM{{\mathcal F}}
\def\TM{{\mathcal T}}
\def\UM{{\mathcal U}}
\def\XM{{\mathcal X}}
\def\YM{{\mathcal Y}}
\def\NM{{\mathcal N}}
\def\OM{{\mathcal O}}
\def\IM{{\mathcal I}}
\def\GM{{\mathcal G}}
\def\PM{{\mathcal P}}
\def\LM{{\mathcal L}}
\def\MM{{\mathcal M}}
\def\DM{{\mathcal D}}
\def\SM{{\mathcal S}}
\def\RB{{\mathbb R}}
\def\EB{{\mathbb E}}

\def\tx{\tilde{\bf x}}
\def\ty{\tilde{\bf y}}
\def\tz{\tilde{\bf z}}
\def\hd{\hat{d}}
\def\HD{\hat{\bf D}}
\def\hx{\hat{\bf x}}
\def\hR{\hat{R}}

\def\Ome{\mbox{\boldmath$\omega$\unboldmath}}
\def\bet{\mbox{\boldmath$\beta$\unboldmath}}
\def\et{\mbox{\boldmath$\eta$\unboldmath}}
\def\ep{\mbox{\boldmath$\epsilon$\unboldmath}}
\def\ph{\mbox{\boldmath$\phi$\unboldmath}}
\def\Pii{\mbox{\boldmath$\Pi$\unboldmath}}
\def\pii{\mbox{\boldmath$\pi$\unboldmath}}
\def\Ph{\mbox{\boldmath$\Phi$\unboldmath}}
\def\Ps{\mbox{\boldmath$\Psi$\unboldmath}}
\def\pss{\mbox{\boldmath$\psi$\unboldmath}}
\def\tha{\mbox{\boldmath$\theta$\unboldmath}}
\def\Tha{\mbox{\boldmath$\Theta$\unboldmath}}
\def\muu{\mbox{\boldmath$\mu$\unboldmath}}
\def\Si{\mbox{\boldmath$\Sigma$\unboldmath}}
\def\Gam{\mbox{\boldmath$\Gamma$\unboldmath}}
\def\gamm{\mbox{\boldmath$\gamma$\unboldmath}}
\def\Lam{\mbox{\boldmath$\Lambda$\unboldmath}}
\def\De{\mbox{\boldmath$\Delta$\unboldmath}}
\def\vps{\mbox{\boldmath$\varepsilon$\unboldmath}}
\def\Up{\mbox{\boldmath$\Upsilon$\unboldmath}}
\def\Lap{\mbox{\boldmath$\LM$\unboldmath}}
\newcommand{\ti}[1]{\tilde{#1}}

\def\tr{\mathrm{tr}}
\def\etr{\mathrm{etr}}
\def\etal{{\em et al.\/}\,}
\newcommand{\indep}{{\;\bot\!\!\!\!\!\!\bot\;}}
\def\argmax{\mathop{\rm argmax}}
\def\argmin{\mathop{\rm argmin}}
\def\vec{\text{vec}}
\def\cov{\text{cov}}
\def\dg{\text{diag}}

\newcommand{\tabref}[1]{Table~\ref{#1}}
\newcommand{\lemref}[1]{Lemma~\ref{#1}}
\newcommand{\thmref}[1]{Theorem~\ref{#1}}
\newcommand{\clmref}[1]{Claim~\ref{#1}}
\newcommand{\crlref}[1]{Corollary~\ref{#1}}
\newcommand{\eqnref}[1]{Eqn.~\ref{#1}}

\newtheorem{remark}{Remark}
\newtheorem{theorem}{Theorem}
\newtheorem{lemma}{Lemma}
\newtheorem{definition}{Definition}

\newtheorem{proposition}{Proposition}

\newcommand{\kexin}[1]{{\color{violet}{(\framebox{Kexin:} #1)}}}

\newcommand{\haoyu}[1]{\textcolor{blue}{\bfseries[Haoyu: #1]}}
\begin{abstract}

Robust perception and reasoning requires consistency across sensory modalities. Yet, current multimodal models often violate this principle, yielding contradictory predictions for visual versus textual representations of the same concept. Rather than masking these failures with standard voting mechanisms—which amplify systematic biases—we demonstrate that cross-modal inconsistency provides a rich, natural signal for learning. We introduce \modelname, a reinforcement learning framework that resolves internal conflicts by enforcing cross-modal cycle consistency. By requiring a model to perform backward inference, switches modalities, and reliably reconstruct the answer via forward inference, we establish a dense, label-free reward. This cyclic constraint forces the model to autonomously align its representations. Optimizing for this structure mitigates modality-specific errors and improves reasoning accuracy by up to 7.6 points. Our results suggest that advanced reasoning emerges not just from scaling data, but from enforcing a structurally consistent understanding of the world.

\end{abstract}    
\section{Introduction}
\label{sec:intro}

\begin{figure}[t]
  \centering
\includegraphics[width=0.49\textwidth]{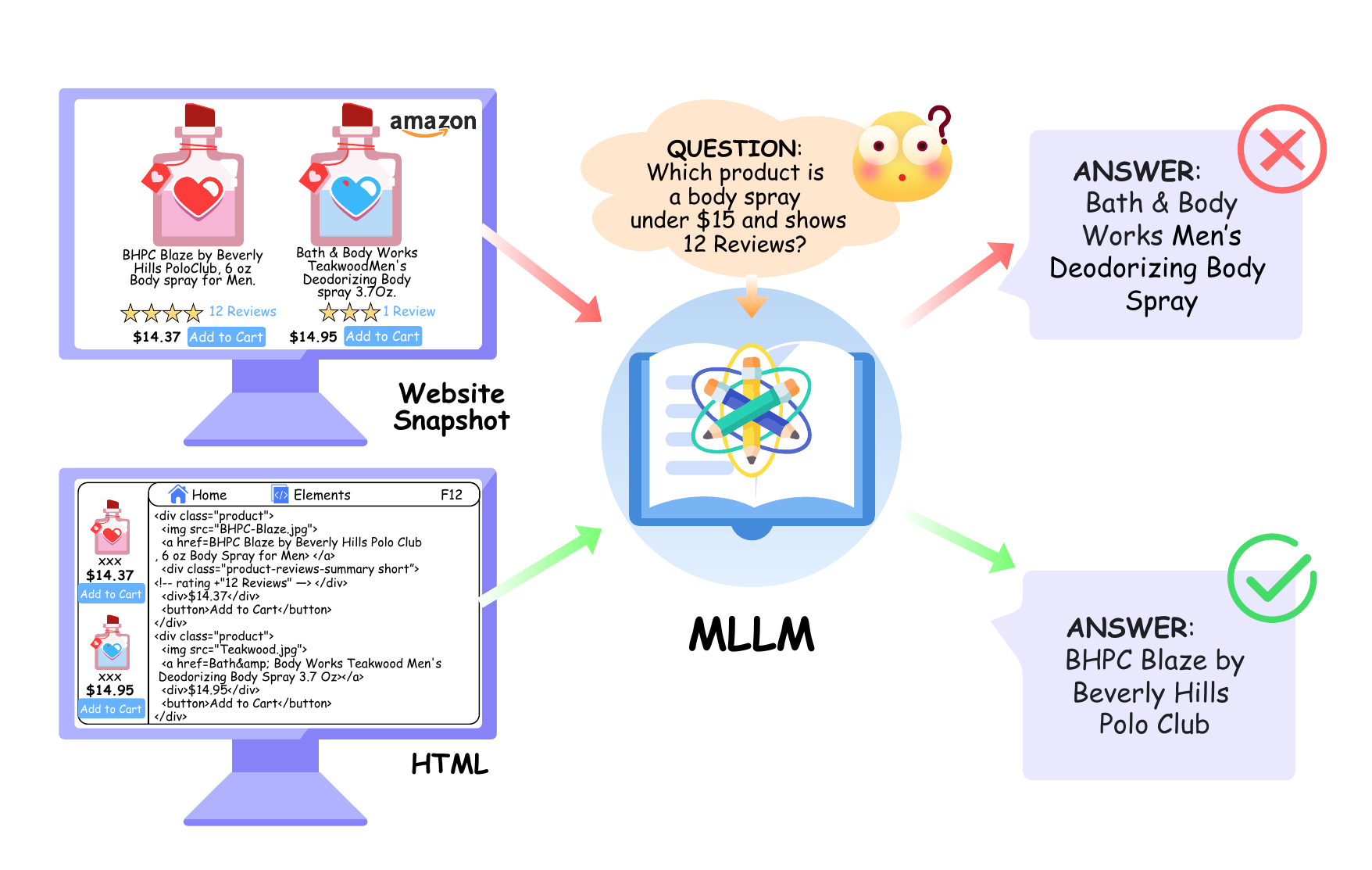}
  \caption{\textbf{Gap in Multimodal Reasoning.} Multimodal large language models (MLLMs) frequently fail the test of modal-invariance. For example, they produce conflicting answers for the same webpage when presented as a screenshot versus its raw HTML source. We introduce a cycle-consistency framework that directly targets this modality gap, leveraging the inconsistency itself as a signal to jointly improve reasoning and alignment.}
   \label{fig:teaser}
\end{figure}

Multimodal Large Language Models (MLLMs) suffer from a fundamental modality gap~\cite{zhu2024unraveling}, often contradicting themselves on visual versus text views of the same input content. As shown in Figure~\ref{fig:teaser}, we show that an MLLM can yield different answers for an identical webpage when presented as a screenshot versus its raw HTML source. As MLLMs are widely deployed in domains such as multimodal document understanding~\cite{mathew2021docvqa,mathew2022infographicvqa}, web UI navigation~\cite{koh2024visualwebarena}, and agentic systems~\cite{mialon2023gaia,yang2023mm,xie2024osworld}, this lack of robustness and consistency can be a critical failure.

\begin{figure*}[t]
  \centering
  \includegraphics[width=\textwidth]{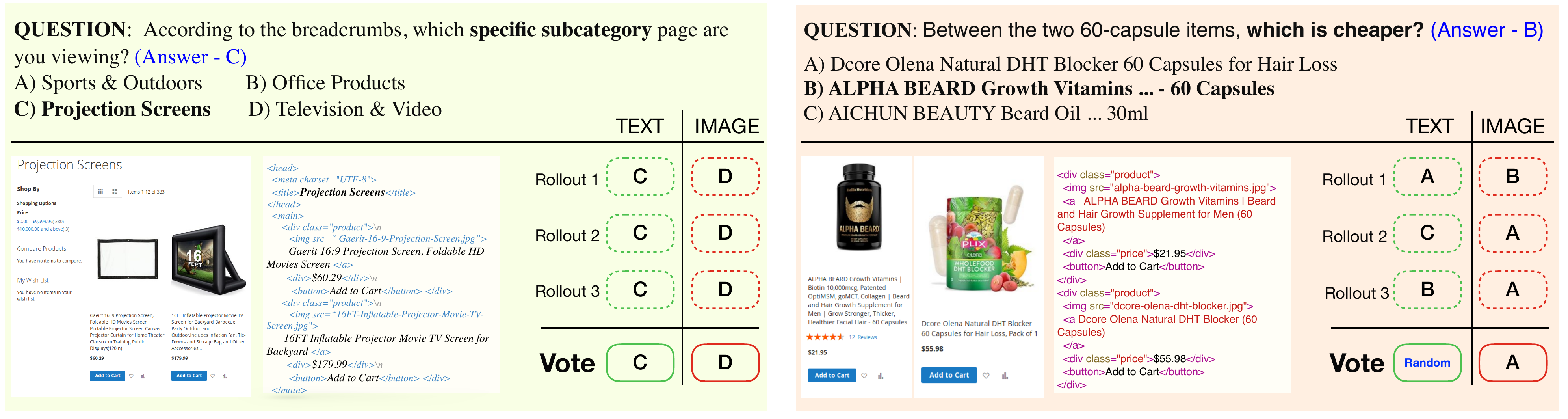}
    \caption{
    \textbf{Failure of multimodal voting.}
    \emph{Left: Consistent Conflict} — both text and image modalities produce self-consistent predictions (\emph{mode-stable}) but disagree with each other, and only one modality aligns with the ground truth.
    \emph{Right: Unstable Recovery} — within a single modality, some rollouts yield the correct answer, but the majority vote remains wrong, reflecting intra-modal instability.
    Using multimodal voting can amplify biases or lose correct signals.
    }
  \label{fig:voting_failure}
\end{figure*}

Most prior work attempts to improve reasoning through large-scale fine-tuning on meticulously curated datasets, which are expensive to construct and inherently limited in their ability to scale model performance~\cite{wang2022self}. Reinforcement learning (RL)~\cite{ouyang2022training,dpo} offers an alternative, but hinges on reliable reward signals; unlike math~\cite{zelikman2024star, wen2025reinforcement} or code~\cite{liu2025code, su2025crossing}, complex multimodal answers are rarely verifiable. 
Without annotated query-answer pairs, recent self-improvement methods use majority voting~\cite{huang2025r,wang2022self1,wang2025ranked}. However, these voting mechanisms suffer from inherent limitations. As illustrated in Figure~\ref{fig:voting_failure}, the problem is compounded in multimodal settings: when visual and textual predictions disagree---an extremely common scenario---the consensus becomes unstable and arbitrary. This leaves the underlying conflict unresolved and, in ``majority-is-wrong'' cases, actually amplifies the error.

\begin{figure*}[t]
  \centering
  \includegraphics[width=0.9\textwidth]{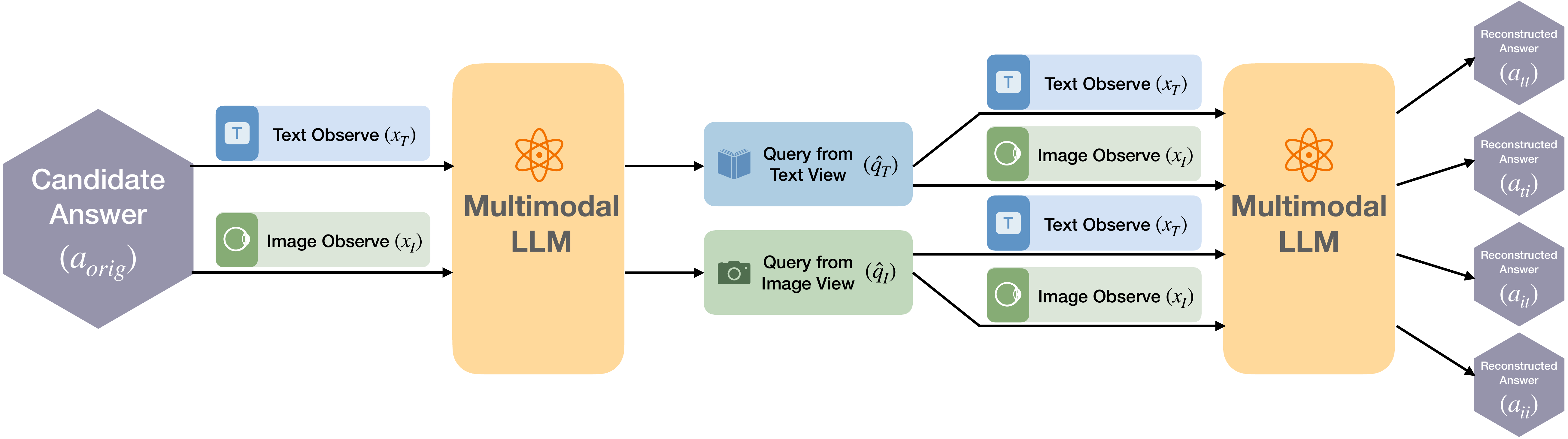}

\caption{
    \textbf{Overview of multimodal cycle consistency.}
    Starting from a potential answer candidate $a_{\text{orig}}$, the model performs \textit{backward inference} to reconstruct
    two latent queries, $\hat{q}_T$ from the text view $x_T$ and $\hat{q}_I$ from the image view $x_I$.
    Each reconstructed query is then used for \textit{forward inference} across both modalities,
    resulting in four reconstructed answers $\{a_{tt}, a_{ti}, a_{it}, a_{ii}\}$ generated via the paths
    $T{\rightarrow}T$, $T{\rightarrow}I$, $I{\rightarrow}T$, and $I{\rightarrow}I$.
    Cycle consistency is measured by whether the reconstructed answers remain consistent with the original
    $a_{\text{orig}}$, forming a full 4-way cross-modal reasoning cycle.
    }
  \label{fig:method}
\end{figure*}

Our key insight is that this cross-modal inconsistency is not a failure but a powerful, untapped resource for self-reward learning.  Instead of relying on flawed voting, we introduce cross-modal cycle consistency (\modelname), a framework that reframes this gap as a self-supervised reward signal. \modelname starts from a candidate answer. Then it performs backward inference to propose a query that would elicit that answer, then switches modalities to perform forward inference and reconstruct the original answer, as shown in Figure~\ref{fig:method}. This cycle serves as a dense, label-free reward that forces the model to resolve its own internal multimodal conflicts, thus improving the reasoning capability.

Extensive experiments, complemented by a diverse suite of case-study visualizations that reveal modality conflicts and how \modelname resolves them, demonstrate that \modelname significantly improves multimodal reasoning capabilities without human annotations. 
In 3B and 8B multimodal LLMs, our method improves performance by up to 7.6 points on major multimodal benchmarks, including ScienceQA~\cite{lu2022learn}, ChartQA~\cite{masry2022chartqa}, InfoVQA~\cite{mathew2022infographicvqa}, MathVista~\cite{lu2023mathvista}, A-OKVQA~\cite{schwenk2022okvqa}, and Visual Web Arena~\cite{koh2024visualwebarena}. Moreover, our method greatly increases the consistency of cross-modal prediction. We further study the conditions under which our cross-modal cycle consistency approach provides the greatest benefit, offering insights into the nature of the modality gap in
state-of-the-art models.

\section{Related Work}

\textbf{Multimodal Large Language Models.}
Vision–language LLMs~\cite{liu2023visual, team2024gemma, yang2025qwen3, lu2024deepseek} extend Language Models (LMs)~\cite{brown2020language, achiam2023gpt, touvron2023llama, bai2023qwen} with a vision encoder that maps images into the token (or embedding) space of the LM. 
In most systems, the vision encoder is pretrained separately and then frozen~\cite{team2024gemma}, and the supervision available for image–text is far sparser than for text-only corpora. These factors contribute to a \emph{modality gap}: the same query can yield different answers depending on whether relevant information is provided as text or embedded in an image. Recent studies begin to characterize this gap~\cite{yuan2025gsm8k, tang2025seam}, but practical methods to close it remain limited. A prevailing strategy is to synthesize additional image–text QA pairs by first captioning images and then prompting an LM to generate query–answer pairs from the captions~\cite{dai2023instructblip, chen2023x, zhang2023llama, wang2023visionllm, liu2023visual, cascante2022simvqa, gupta2022swapmix}; however, such pipelines often require nontrivial human curation and can propagate caption biases~\cite{doveh2023dense}. 
In contrast, we leverage incidental structure in naturally occurring multimodal data to improve understanding and reasoning without relying on large volumes of manually curated synthetic QA.


\noindent\textbf{Reward Modeling.}
Reinforcement learning from human feedback (RLHF) and its extensions have been central to aligning LLMs with human preferences~\cite{ouyang2022training, guo2025deepseek}. For verifiable domains such as mathematics and code~\cite{lambert2024tulu}, outcome-based rewards are sufficient because correctness can be objectively measured. Beyond outcome-only supervision~\cite{shao2024deepseekmath}, recent work explores rewarding intermediate reasoning steps~\cite{wei2022chain,feng2023towards, xu2025direct}, often pairing process supervision with learned reward models that estimate the quality of partial solutions rather than final answers~\cite{li2025semantically, ma2025general, li2024fg, wen2025reinforcement}. Although these methods are effective in domains with verifiable feedback, they remain brittle in multimodal reasoning, where step-level evaluation is ambiguous, and learned reward models can inherit modality-specific biases. This leads to \emph{reward misspecification}—overfitting to superficial textual or visual cues rather than genuine semantic correctness.

\noindent\textbf{Label-Free Reinforcement Learning and Self-Evolution.}
To mitigate reliance on human or synthetic labels, recent trends pursue \emph{label-free} self-improvement paradigms, including label-free RL~\cite{huang2025r}, self-play~\cite{tu2024towards}, self-instruct~\cite{wang2022self}, and self-training or self-refinement~\cite{gulcehre2023reinforced,huang2022large,zhuge2024agent}. These methods iteratively generate and refine their own data using internal feedback, sometimes leveraging consistency or confidence as reward signals~\cite{li2025semantically,zuo2025ttrl,zhang2025consistent}. More advanced frameworks extend this idea to weak-to-strong learning~\cite{leike2023superalignment,burns2023weak} and self-critique mechanisms~\cite{chen2024self,yuan2024self,chen2025spc}, where models bootstrap improvement from self-generated critics. However, many systems still equate consensus with correctness, optimizing for agreement through majority voting~\cite{chen2025pass}. This risks amplifying systematic biases, a phenomenon often described as a “majority-is-wrong” failure.
In multimodal LLMs, this issue is exacerbated by \emph{cross-modality inconsistency}~\cite{zhu2024unraveling}, where visual and textual predictions diverge even when grounded in the same underlying content. Our method avoids voting through a cycle consistency.

\section{Method}

Our goal is to improve the multimodal reasoning capabilities of MLLM. We frame this as a reinforcement learning (RL) problem, where the key challenge is to acquire a reward signal without human labels. We first review self-rewarding methods based on consensus voting, showing how they fail in single-modal settings and how this failure is compounded in multimodal contexts. We then introduce our solution, \modelname, a self-supervised reward framework that replaces flawed consensus with cross-modal cycle consistency.

\subsection{The Failure of Consensus-Based Rewards}
\label{sec:consensus_failure}

\begin{figure*}[t]
  \centering
\includegraphics[width=1.0\textwidth]{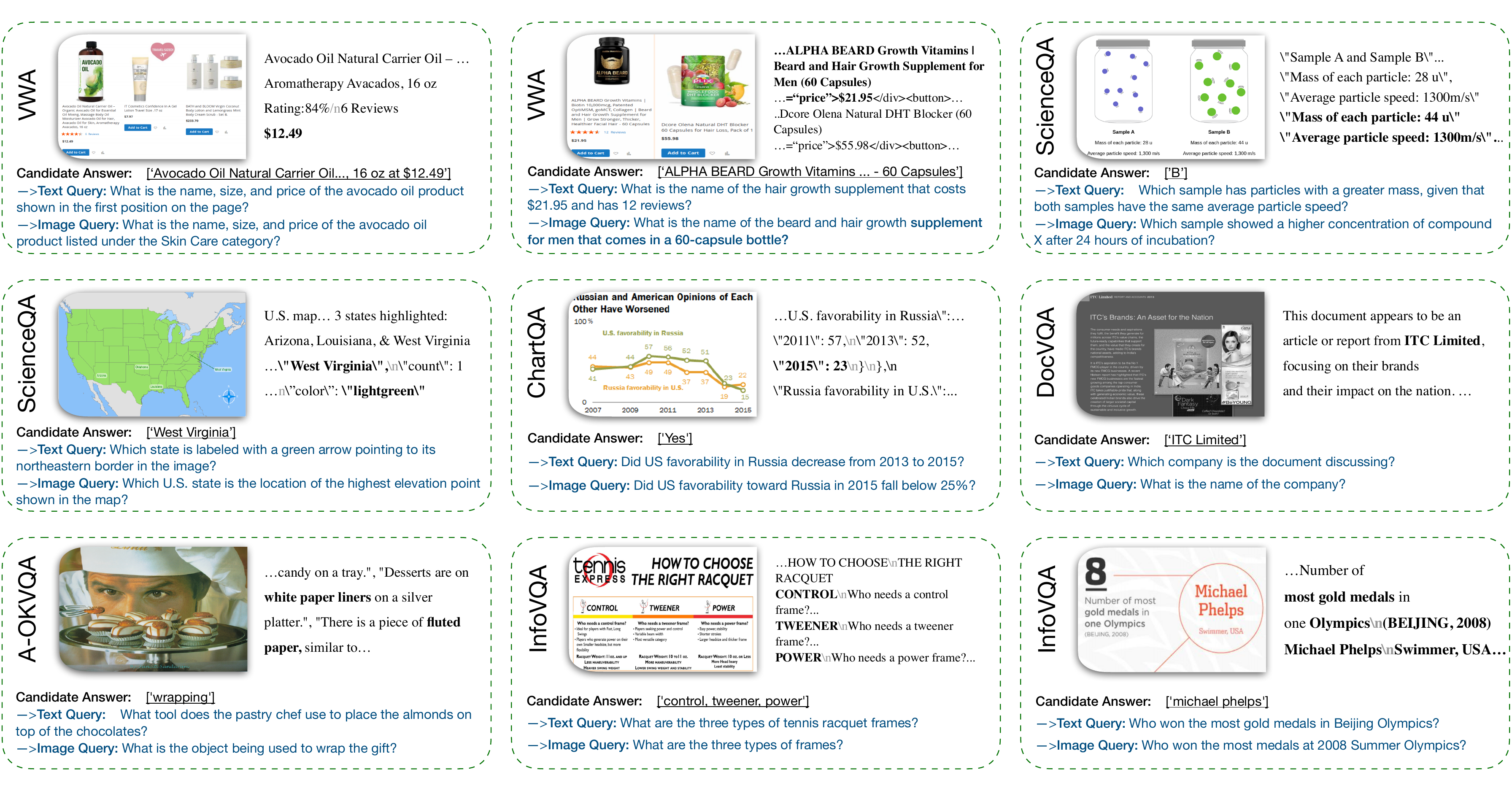}
 \caption{\textbf{Examples of the backward-inference (Answer$\to$Query) step}. Given a {Candidate Answer}, the model generates distinct, semantically-grounded queries for both the text and image modalities. This demonstrates the viability of the first step of our cycle-consistency reward, enabling the model to check its answer in the alternate modality.}
   \label{fig:cycle-examples}
\end{figure*}

We define our MLLM as a model $F_\theta$ that produces an answer $\hat{\a}$ given a multimodal input $\mathbf{x}$ and a text query $\q$: $\hat{\a} = F_\theta(\mathbf{x}, \q)$. The input $\mathbf{x}$ can be visual-only $\mathbf{x}_I$ (e.g., a screenshot), text-only $\mathbf{x}_T$ (e.g., HTML), or an interleaved mix $\mathbf{x}_M$.


To improve MLLM's reasoning capability, we follow the standard practice and optimize the model with reinforcement learning, a GRPO objective~\cite{shao2024deepseekmath}. At each step, $F_{\theta}$ generates an answer for the query; a reward model assigns a scalar score, and we update $\theta$ to increase the likelihood of the sampled answer when the reward is high and decrease it when the reward is low. This aligns $F_{\theta}$ towards responses that consistently earn a higher reward.
Formally, we will maximize this objective:

$$\mathcal{L}_{\text{GRPO}} = \mathbb{E} \left[ \log \pi_\theta(\hat{\a}_i | \mathbf{x}_i, \q_i) \cdot \hat{A}(\hat{\a_i}, \a_i) \right]$$
where $A(\hat{a})$ is the advantage, which is calculated via
\begin{equation}
\hat{A}(\hat{\a_i}, \a_i) = \frac{r_i - \mathrm{mean}(\mathbf{r})}{\mathrm{std}(\mathbf{r})},
\end{equation}
where the advantage is calculated on the basis of its relative normalized value in the batch.
This paradigm, however, hinges on a reliable reward function $r_i = R(\hat{\a_i}, \a_i)$ that requires a ground-truth answer $\a$ or an external verifier, both of which can be expensive to scale for many multimodal tasks.

\noindent\textbf{Single-Modal Voting and ``Majority-is-Wrong''.} To remove the need for labels, recent ``self-improvement'' methods like R-zero~\cite{huang2025r} generate their own rewards without labeled query-answer pairs. 
For a given text query, the model generates $k$ candidate answers $\{\a_j\}_{j=1}^k$. Then it applies majority voting to select the most plausible answer $\a'$, which serves as a pseudo-label. 
\[
    \a' = \mathrm{mode}(\{\a_j\}_{j=1}^k).
\]
It calculates the reward as $r_i = R(\hat{\a_i}, \a')$. The model is then trained to increase the probability of producing this majority-voted answer.

This approach suffers from a fundamental ``majority-is-wrong'' failure. 
If the model has a systematic bias, the majority of its answers will be incorrect. 
Voting will select an incorrect pseudo-label, and the RL objective will simply reinforce the model's own mistakes, leading to a collapse in performance.

\noindent\textbf{Compounded Failure: Multimodal Voting.}  
To extend voting to multimodal reasoning, we aggregate predictions from both the image and text modalities.  
For each query, the model answers $k_I$ times from image views and $k_T$ times from text views, predictions from the text modality $\{\a_t^{T}\}$ and the image modality $\{\a_j^{I}\}$ are pooled together, and the pseudo-label is determined by majority vote:

\[
\a'_{\text{multi}} = 
\mathrm{mode}\big( \{\a_j^{I}\}_{j=1}^{k_I} \cup \{\a_t^{T}\}_{t=1}^{k_T} \big).
\]

The reward is then computed in the same way as the single-modal case,  
$r_i = R(\hat{\a}_i, \a'_{\text{multi}})$.

However, the problem of voting is compounded when applied to multimodal reasoning. A naive extension is to generate answers from different modalities and vote. However, we find that predictions from different modalities frequently disagree. As shown in Figure~\ref{fig:voting_failure}, when the model is presented with the same information in different forms (e.g., a chart image $\mathbf{x}_I$ vs. its data table $\mathbf{x}_T$), its answers often conflict. This cross-modal disagreement makes a simple majority vote unstable or arbitrary. If the visual and textual predictions differ, there is no clear consensus to follow, leaving the model without a reliable learning signal.

\subsection{R-C$^2$: Self-Reward from Cycle Consistency}
To break this cycle of self-reinforcing errors, we propose the Cross-Modal Cycle Consistency Reward (\modelname). Our framework replaces flawed, unstable voting with a dense, self-supervised reward signal derived from the model's own internal consistency.

Our key insight is to shift from answer-side voting to answer-side verification. Instead of starting from a query to aggregate answers, we begin with a single candidate answer $\a_{\text{orig}}$—which may come from the model’s own prediction or any available answer source, without relying on query–answer labels—and assess its logical self-consistency. 
To achieve this, MLLM first \textbf{backward infers} the probable queries that could elicit this answer $\a_{\text{orig}}$.
This is achieved by conditioning on both $\a_{\text{orig}}$ and the content of one modality (e.g., text view $\x_T$), and prompting the MLLM to predict.  This inferred query $\hat{q}$ is then fed back into the model, but this time conditioned on the \textit{alternate} modality (e.g., image view $\x_I$), to \textbf{forward infer} a reconstructed answer, $\hat{\a}$.
See Figure~\ref{fig:cycle-examples} for examples of the cycle consistency data. A binary reward is generated if the reconstructed $\hat{\a}$ is consistent with the original $\a_{\text{orig}}$:
\[
r =
\begin{cases}
1, & \text{if } \hat{\a} \text{  matches } \a_{\text{orig}},\\
0, & \text{otherwise.}
\end{cases}
\]
This self-verification design provides a reliable, self-supervised reward, eliminating the need for labeled query-answer pairs.

\paragraph{Multimodal Cycle Consistency.}
As shown in Figure~\ref{fig:method},
\modelname evaluates whether a candidate answer $\a_{\text{orig}}$ is stable by running 
a complete set of backward–forward cycles in both modalities.  Given a sample with synchronized textual ($\x_T$) and visual ($\x_I$) views, we leverage both to form four distinct reasoning paths.

First, in the \textbf{backward step}, the model infers two types of queries from $\a_{\text{orig}}$: $\hat{q}_T$ (conditioned on $\x_T$) and $\hat{q}_I$ (conditioned on $\x_I$).
Next, in the \textbf{forward step}, each query is verified against both modalities. 

This results in four complete cycle directions:
\[
T{\rightarrow}T,\quad T{\rightarrow}I,\quad I{\rightarrow}T,\quad I{\rightarrow}I.
\]
A reward is generated for each path where every $a \in \{a_{tt}, a_{ii}, a_{ti}, a_{it}\}$ matches $\a_{\text{orig}}$. Crucially, \modelname\ uses all four cycles. Same-modality cycles ($T{\rightarrow}T$, $I{\rightarrow}I$) enforce internal stability. Cross-modality cycles ($T{\rightarrow}I$, $I{\rightarrow}T$) force the model to \textbf{resolve its modality gap} and align semantics. This full 4-way evaluation provides a comprehensive signal, demanding both intra-modal robustness and cross-modal agreement.



\paragraph{Training Pipeline.}
While we can do this cross-modal cycle consistency online, which means we dynamically generate new cycle data using the model updated from the previous step, we adopt an offline strategy, which pre-generates the entire synthetic cycle dataset before training begins. While the online method allows the data to co-evolve with the model, we find the offline version is substantially more training-efficient, as it permits the pre-computation and batching of all cycle data. Our full pipeline proceeds in three stages: (0) an optional step for multimodal data preparation, followed by our core two-stage algorithm: (1) constructing cycles from candidate answers and (2) reinforcement learning with cycle rewards.

\noindent \textit{(0) Multimodal data preparation.}  
Each training sample provides semantically aligned visual and textual views.  
For datasets that naturally contain both modalities (e.g., webpages and their corresponding HTML), we directly use the paired image-text inputs.  
For image-only datasets, we obtain the textual view by prompting the MLLM we use to generate a reliable semantic description.  
This ensures that every sample contains a coherent multimodal representation suitable for cross-modal cycles.

\noindent \textit{(1) Constructing cycles from candidate answers.}  
Training begins by sampling one or more \emph{candidate answers} from the model’s current policy using either the image or the text modality.  
Each candidate answer is then used to infer a backward query through the same MLLM model.  
The resulting $(\hat{q}, a_{\text{orig}})$ pair forms a synthetic query–answer pair that reflects the model’s internal reasoning.  
These pairs serve as the data for cycle verification and provide self-generated supervision without requiring labeled QA data.

\noindent \textit{(2) Reinforcement learning with cycle rewards.}  
Given the backward query, the model performs a forward pass (in the same or alternate modality) to obtain a reconstructed answer.  
Comparing this answer with the original one yields the binary cycle-consistency reward for GRPO training.

\section{Experiments}

\begin{table*}[t]
\centering
\small
\renewcommand{\arraystretch}{1.05}
\setlength{\tabcolsep}{5pt}

\begin{tabular}{l|cc|cc|cc|>{\columncolor[gray]{0.9}}c>{\columncolor[gray]{0.9}}c}
\toprule
 & \multicolumn{8}{c}{\textbf{Qwen2.5-VL-3B-Instruct}} \\
 \midrule
\multirow{2}{*}{\textbf{Dataset}} 
& \multicolumn{2}{c|}{\textbf{Base Model}} 
& \multicolumn{2}{c|}{\textbf{+ Voting (Text)}} 
& \multicolumn{2}{c|}{\textbf{+ Voting (Image+Text)}} 
& \multicolumn{2}{c}{\textbf{+ \modelname (Ours)}} \\
\cmidrule(lr){2-9}
 & Text Acc & Vision Acc
 & Text Acc & Vision Acc
 & Text Acc & Vision Acc
 & Text Acc & Vision Acc \\
\midrule
\textbf{ScienceQA}   & 68.9 & 76.0 & 70.7 {\textcolor{gray!90}{(+1.8)}}   & 76.2 {\textcolor{gray!90}{(+0.2)}}   & 73.1 {\textcolor{gray!90}{(+4.2)}}   & 78.0 {\textcolor{gray!90}{(+2.0)}}   & \textbf{76.7 {\color[HTML]{006400}(+7.8)}} & \textbf{83.3 {\color[HTML]{006400}(+7.3)}} \\
\textbf{ChartQA}     & 71.1 & 82.8 & 76.0 {\textcolor{gray!90}{(+4.9)}}   & 83.6 {\textcolor{gray!90}{(+0.8)}}   & 76.2 {\textcolor{gray!90}{(+5.1)}}   & 83.5 {\textcolor{gray!90}{(+0.7)}}   & \textbf{77.2 {\color[HTML]{006400}(+6.1)}} & \textbf{84.8 {\color[HTML]{006400}(+2.0)}} \\
\textbf{MathVista}   & 49.8 & 64.8 & 50.6 {\textcolor{gray!90}{(+0.8)}}   & 65.0 {\textcolor{gray!90}{(+0.2)}}   & 52.1 {\textcolor{gray!90}{(+2.3)}}   & 65.7 {\textcolor{gray!90}{(+0.9)}}   & \textbf{55.8 {\color[HTML]{006400}(+6.0)}} & \textbf{67.6 {\color[HTML]{006400}(+2.8)}} \\
\textbf{VWA}         & 69.0 & 62.9 & \textbf{74.5 {\textcolor{gray!90}{(+5.5)}}} & 63.1 {\textcolor{gray!90}{(+0.2)}}   & 73.3 {\textcolor{gray!90}{(+4.3)}}   & 64.5 {\textcolor{gray!90}{(+1.6)}}   & \textbf{74.5 {\color[HTML]{006400}(+5.5)}} & \textbf{67.1 {\color[HTML]{006400}(+4.2)}} \\
\textbf{A-OKVQA}      & 70.3 & 86.1 & 73.9 {\textcolor{gray!90}{(+3.6)}}   & 87.6 {\textcolor{gray!90}{(+1.5)}}   & 74.2 {\textcolor{gray!90}{(+3.9)}}   & 87.7 {\textcolor{gray!90}{(+1.6)}}   & \textbf{75.2} {\textbf{\color[HTML]{006400}(+4.9)}} & \textbf{88.8} {\textbf{\color[HTML]{006400}(+2.7)}} \\
\textbf{DocVQA}      & 74.7 & 90.0 & 76.2 {\textcolor{gray!90}{(+1.5)}}   & 90.3 {\textcolor{gray!90}{(+0.3)}}   & 76.4 {\textcolor{gray!90}{(+1.7)}}   & \textbf{90.4} {\textcolor{gray!90}{\textbf{(+0.4)}}}   & \textbf{76.6} {\textbf{\color[HTML]{006400}(+1.9)}} & 90.3 {\color[HTML]{006400}(+0.3)} \\
\textbf{InfoVQA}     & 54.9 & 74.1 & 55.6 {\textcolor{gray!90}{(+0.7)}}   & 74.2 {\textcolor{gray!90}{(+0.1)}}   & 56.1 {\textcolor{gray!90}{(+1.2)}}   & \textbf{74.3 {\textcolor{gray!90}{(+0.2)}}} & \textbf{56.3 {\color[HTML]{006400}(+1.4)}}  & \textbf{74.3 {\color[HTML]{006400}(+0.2)}}  \\
\midrule
\textbf{Average}  & 65.5 & 76.7 & 68.2 {\textcolor{gray!90}{(+2.7)}} & 77.1 {\textcolor{gray!90}{(+0.4)}} & 68.8 {\textcolor{gray!90}{(+3.3)}} & 77.7 {\textcolor{gray!90}{(+1.0)}} & \textbf{70.3 {\color[HTML]{006400}(+4.8)}} & \textbf{79.5 {\color[HTML]{006400}(+2.8)}} \\
\bottomrule


\toprule
 & \multicolumn{8}{c}{\textbf{Qwen3-VL-8B-Instruct}} \\
 \midrule
\multirow{2}{*}{\textbf{Dataset}} 
& \multicolumn{2}{c|}{\textbf{Base Model}} 
& \multicolumn{2}{c|}{\textbf{+ Voting (Text)}} 
& \multicolumn{2}{c|}{\textbf{+ Voting (Image+Text)}} 
& \multicolumn{2}{c}{\textbf{+ \modelname (Ours)}} \\
\cmidrule(lr){2-9}
 & Text Acc & Vision Acc
 & Text Acc & Vision Acc
 & Text Acc & Vision Acc
 & Text Acc & Vision Acc \\
\midrule
\textbf{ScienceQA}   & 72.9 & 90.3 & 72.6 {\textcolor{gray!90}{(-0.3)}} & 90.0 {\textcolor{gray!90}{(-0.3)}} & 74.8 {\textcolor{gray!90}{(+1.9)}} & 90.0 {\textcolor{gray!90}{(-0.3)}} & \textbf{75.9 {\color[HTML]{006400}(+3.0)}} & \textbf{91.3 {\color[HTML]{006400}(+1.0)}} \\
\textbf{ChartQA}     & 78.8 & 85.1 & 80.2 {\textcolor{gray!90}{(+1.4)}} & 85.3 {\textcolor{gray!90}{(+0.2)}} & 80.2 {\textcolor{gray!90}{(+1.4)}} & 85.8 {\textcolor{gray!90}{(+0.7)}} & \textbf{80.8 {\color[HTML]{006400}(+2.0)}} & \textbf{86.0 {\color[HTML]{006400}(+0.9)}} \\
\textbf{MathVista}   & 62.0 & 73.8 & 61.8 {\textcolor{gray!90}{(-0.2)}} & 73.7 {\textcolor{gray!90}{(-0.1)}} & 64.8 {\textcolor{gray!90}{(+2.8)}} & 76.5 {\textcolor{gray!90}{(+2.7)}} & \textbf{66.3 {\color[HTML]{006400}(+4.3)}} & \textbf{76.9 {\color[HTML]{006400}(+3.1)}} \\
\textbf{VWA}         & 84.3 & 81.2 & 85.1 {\textcolor{gray!90}{(+0.8)}} & 81.6 {\textcolor{gray!90}{(+0.4)}} & 85.3 {\textcolor{gray!90}{(+1.0)}} & 81.0 {\textcolor{gray!90}{(-0.2)}} & \textbf{85.7 {\color[HTML]{006400}(+1.4)}} & \textbf{81.8 {\color[HTML]{006400}(+0.6)}} \\
\textbf{A-OKVQA}      & 72.2 & 86.4 & 72.8 {\textcolor{gray!90}{(+0.6)}}   & 86.8 {\textcolor{gray!90}{(+0.4)}}   & 73.4 {\textcolor{gray!90}{(+1.2)}}   & 86.9 {\textcolor{gray!90}{\textbf{(+0.5)}}}   & \textbf{73.5} {\textbf{\color[HTML]{006400}(+1.3)}} & \textbf{87.3 {\color[HTML]{006400}(+0.9)}} \\
\textbf{DocVQA}      & 78.0 & 92.9 & 79.2 {\textcolor{gray!90}{(+1.2)}} & 93.6 {\textcolor{gray!90}{(+0.7)}} & \textbf{79.4 {\textcolor{gray!90}{(+1.4)}}} & \textbf{93.8 {\textcolor{gray!90}{(+0.9)}}} & \textbf{79.4 {\color[HTML]{006400}(+1.4)}} & 93.3 {\color[HTML]{006400}(+0.4)} \\
\textbf{InfoVQA}     & 60.4 & 82.6 & 60.7 {\textcolor{gray!90}{(+0.3)}} & 82.5 {\textcolor{gray!90}{(-0.1)}} & 61.9 {\textcolor{gray!90}{(+1.5)}} & 83.1 {\textcolor{gray!90}{(+0.5)}} & \textbf{62.7 {\color[HTML]{006400}(+2.3)}} & \textbf{83.3 {\color[HTML]{006400}(+0.7)}} \\
\midrule
\textbf{Average}  & 72.7 & 84.6 & 73.2 {\textcolor{gray!90}{(+0.5)}} & 84.8 {\textcolor{gray!90}{(+0.2)}} & 74.3 {\textcolor{gray!90}{(+1.6)}} & 85.3 {\textcolor{gray!90}{(+0.7)}} & \textbf{74.9 {\color[HTML]{006400}(+2.2)}} & \textbf{85.7 {\color[HTML]{006400}(+1.1)}} \\
\bottomrule
\end{tabular}


\caption{\textbf{Multimodal Reasoning Accuracy.}
We compare the Base Model, Voting (Text-only), Voting (Image+Text), and our \modelname-RL method on Qwen2.5-VL-3B-Instruct (top) and Qwen3-VL-8B-Instruct (bottom). 
Each cell reports Text Accuracy (input content in text format) and Vision Accuracy (input content in image format), with green numbers indicating absolute improvements of \modelname over the base model. We \textbf{bold} the best accuracy for each dataset.
Multimodal voting consistently outperforms text-only voting and the base model, and \modelname achieves the largest gains across nearly all benchmarks.
}
\label{tab:main_acc}
\end{table*}

\begin{table*}[t]
\centering
\small
\renewcommand{\arraystretch}{1.05}
\setlength{\tabcolsep}{5pt}
\resizebox{\textwidth}{!}{%
\begin{tabular}{l|ccc>{\columncolor[gray]{0.9}}c|ccc>{\columncolor[gray]{0.9}}c}
\toprule
\multirow{2}{*}{\textbf{Dataset}} 
& \multicolumn{4}{c|}{\textbf{Qwen2.5-VL-3B}} 
& \multicolumn{4}{c}{\textbf{Qwen3-VL-8B}} \\
\cmidrule(lr){2-9}
 & Base & Voting (Text) & Voting (Image+Text) & \modelname (Ours) 
 & Base & Voting (Text) & Voting (Image+Text) & \modelname (Ours)  \\
\midrule
\textbf{ScienceQA}   & 74.9 & 79.7 {\textcolor{gray!90}{(+4.8)}} & 81.7 {\textcolor{gray!90}{(+6.8)}} & \textbf{84.9 {\color[HTML]{006400}(+10.0)}} 
                     & 74.7 & 75.3 {\textcolor{gray!90}{(+0.6)}} & 77.3 {\textcolor{gray!90}{(+2.6)}} & \textbf{78.3 {\color[HTML]{006400}(+3.6)}} \\
\textbf{ChartQA}     & 73.4 & 79.2 {\textcolor{gray!90}{(+5.8)}}   & 78.9 {\textcolor{gray!90}{(+5.5)}}   & \textbf{79.5 {\color[HTML]{006400}(+6.1)}} 
                     & 81.4 & 82.7 {\textcolor{gray!90}{(+1.3)}} & 83.2 {\textcolor{gray!90}{(+1.8)}} & \textbf{83.5 {\color[HTML]{006400}(+2.1)}} \\
\textbf{MathVista}   & 57.3 & 58.8 {\textcolor{gray!90}{(+1.5)}} & 59.2 {\textcolor{gray!90}{(+1.9)}} & \textbf{61.4 {\color[HTML]{006400}(+4.1)}} 
                     & 64.7 & 64.8 {\textcolor{gray!90}{(+0.1)}} & 69.0 {\textcolor{gray!90}{(+4.3)}} & \textbf{69.2 {\color[HTML]{006400}(+4.5)}} \\
\textbf{VWA}         & 63.9 & 66.9 {\textcolor{gray!90}{(+3.0)}} & \textbf{70.0 {\textcolor{gray!90}{(+6.1)}}} & 67.1 {\color[HTML]{006400}(+3.2)}  
                     & 79.2 & 80.0 {\textcolor{gray!90}{(+0.8)}} & 79.8 {\textcolor{gray!90}{(+0.6)}} & \textbf{81.2 {\color[HTML]{006400}(+2.0)}} \\
\textbf{A-OKVQA}      & 67.8 & 80.1 {\textcolor{gray!90}{(+12.3)}} & 80.2 {\textcolor{gray!90}{(+12.4)}} & \textbf{80.3 {\color[HTML]{006400}(+12.5)}}    
                     & 70.2 & 79.3 {\textcolor{gray!90}{(+9.1)}} & \textbf{\textbf{80.3 {\textcolor{gray!90}{(+10.1)}}}} & 79.6 {\color[HTML]{006400}(+9.4)} \\
\textbf{DocVQA}      & 75.2 & 76.9 {\textcolor{gray!90}{(+1.7)}} & \textbf{77.1 {\textcolor{gray!90}{(+1.9)}}} & 76.6 {\color[HTML]{006400}(+1.4)}    
                     & 79.2 & 79.9 {\textcolor{gray!90}{(+0.7)}} & \textbf{80.0 {\textcolor{gray!90}{(+0.8)}}} & \textbf{80.0 {\color[HTML]{006400}(+0.8)}} \\
\textbf{InfoVQA}     & 56.5 & 56.9 {\textcolor{gray!90}{(+0.4)}} & 57.8 {\textcolor{gray!90}{(+1.3)}} & \textbf{58.5 {\color[HTML]{006400}(+2.0)}}  
                     & 60.9 & 61.1 {\textcolor{gray!90}{(+0.2)}} & 62.1 {\textcolor{gray!90}{(+1.2)}} & \textbf{65.8 {\color[HTML]{006400}(+4.9)}} \\
\midrule
\textbf{Average}     & 67.0 & 71.2 {\textcolor{gray!90}{(+4.2)}} & 72.1 {\textcolor{gray!90}{(+5.1)}} & \textbf{72.6 {\color[HTML]{006400}(+5.6)}}  
                     & 72.9 & 74.7 {\textcolor{gray!90}{(+1.8)}} & 76.0 {\textcolor{gray!90}{(+3.1)}} & \textbf{76.8{\color[HTML]{006400}(+3.9)}}  \\
\bottomrule
\end{tabular}%
}

\caption{
\textbf{Multimodal Consistency Ratio.} We compare both {Qwen2.5-VL-3B} and {Qwen3-VL-8B}.  
Each dataset is evaluated under four supervision settings:  
{Base}, {+Voting (Text)}, {+Voting (Image+Text)}, and {+\modelname}.  
Values in {\color[HTML]{006400}(green)} indicate absolute improvements of \modelname over the base model. Across most datasets, \modelname achieves the highest cross-modal agreement.
}
\label{tab:main_consistency}
\end{table*}
\subsection{Baselines}
We use Qwen2.5-VL-3B-Instruct and Qwen3-VL-8B-Instruct as the baseline. We also adopt R0's~\cite{huang2025r} majority voting as our baseline: for each query, the model generates multiple rollouts and adopts the most frequent answer as the pseudo-label. In the text-only case, voting is applied over textual rollouts, while in the multimodal setting, predictions from image and text rollouts are pooled together and voted jointly. Unless otherwise stated, we set the number of rollouts to $4$ per modality in all voting-based experiments. See Sec.~\ref{sec:consensus_failure} for procedural details.

\subsection{Implementation Details}
\label{subsec:implementation_details}
We train all models using the GRPO objective with a learning rate of $1e-6$. 
Experiments are conducted on 4 Blackwell 6000 Pro GPUs with mixed-precision training. 
For each update step, the model samples four rollouts per modality using a temperature of 1.0 and a top-$p$ of 0.95. 
We use an effective batch size of 256 for VWA, InfoVQA, and DocVQA (long-input datasets), and 1024 for all other datasets, via gradient accumulation. Larger batch sizes are adopted when GPU memory permits. Training typically runs for 100–150 steps, and we use a validation set for model selection and early stopping.

\subsection{Datasets}
We evaluate \modelname across six multimodal reasoning benchmarks spanning diverse visual–textual understanding tasks.  
\textit{ScienceQA}~\cite{lu2022learn} covers general visual question answering;  
\textit{ChartQA}~\cite{masry2022chartqa} focuses on reasoning over data visualizations;  
\textit{DocVQA}~\cite{mathew2021docvqa}, \textit{InfoVQA}~\cite{mathew2022infographicvqa} emphasize OCR-based reasoning in infographic-style images; \textit{A-OKVQA}~\cite{schwenk2022okvqa} extends VQA into open-domain reasoning over natural images, requiring commonsense and factual knowledge beyond visual perception;
and \textit{MathVista}~\cite{lu2023mathvista} evaluates visual mathematical reasoning grounded in figures and quantitative relations.
We additionally repurpose the \textit{VWA} benchmark~\cite{koh2024visualwebarena} to assess reasoning in complex web environments.  
Specifically, we use its \emph{shopping website} subset and use multiple-choice question–answer pairs to enable consistent and straightforward evaluation.

\begin{table}[t]
\centering
\small
\setlength{\tabcolsep}{5pt}
\renewcommand{\arraystretch}{1.1}
\resizebox{\columnwidth}{!}{
\begin{tabular}{l|ccc|ccc}
\toprule
\multirow{2}{*}{\textbf{Cycle Type}} 
& \multicolumn{3}{c|}{\textbf{ScienceQA}} 
& \multicolumn{3}{c}{\textbf{ChartQA}} \\
\cmidrule(lr){2-7}
 & Text & Vision & Cons. 
 & Text & Vision & Cons. \\
\midrule
\textbf{Single} 
& 74.0 & 81.7 & 81.2
& 76.2 & 83.7 & 77.3\\
\textbf{Cross}
& 75.8 & 80.1 & 83.1
& 76.5 & 83.8 & 78.4\\
\rowcolor{gray!20}
\textbf{Mixed}
& 76.7 & 83.3 & 84.9
& 77.2 & 84.8 & 79.5\\
\bottomrule
\end{tabular}
}
\caption{
Ablation of cycle path configurations in \modelname on ScienceQA and ChartQA using Qwen2.5-VL-3B-Instruct.  
\textit{Single} uses intra-modal cycles (I→I, T→T);  
\textit{Cross} uses cross-modal cycles (I→T, T→I);  
\textit{Mixed} combines all four paths.  
The mixed configuration yields the highest accuracy and consistency.
}
\label{tab:cycle_path_interaction_ablation}
\end{table}

\begin{figure*}[t]
  \centering
\includegraphics[width=1.0\textwidth]{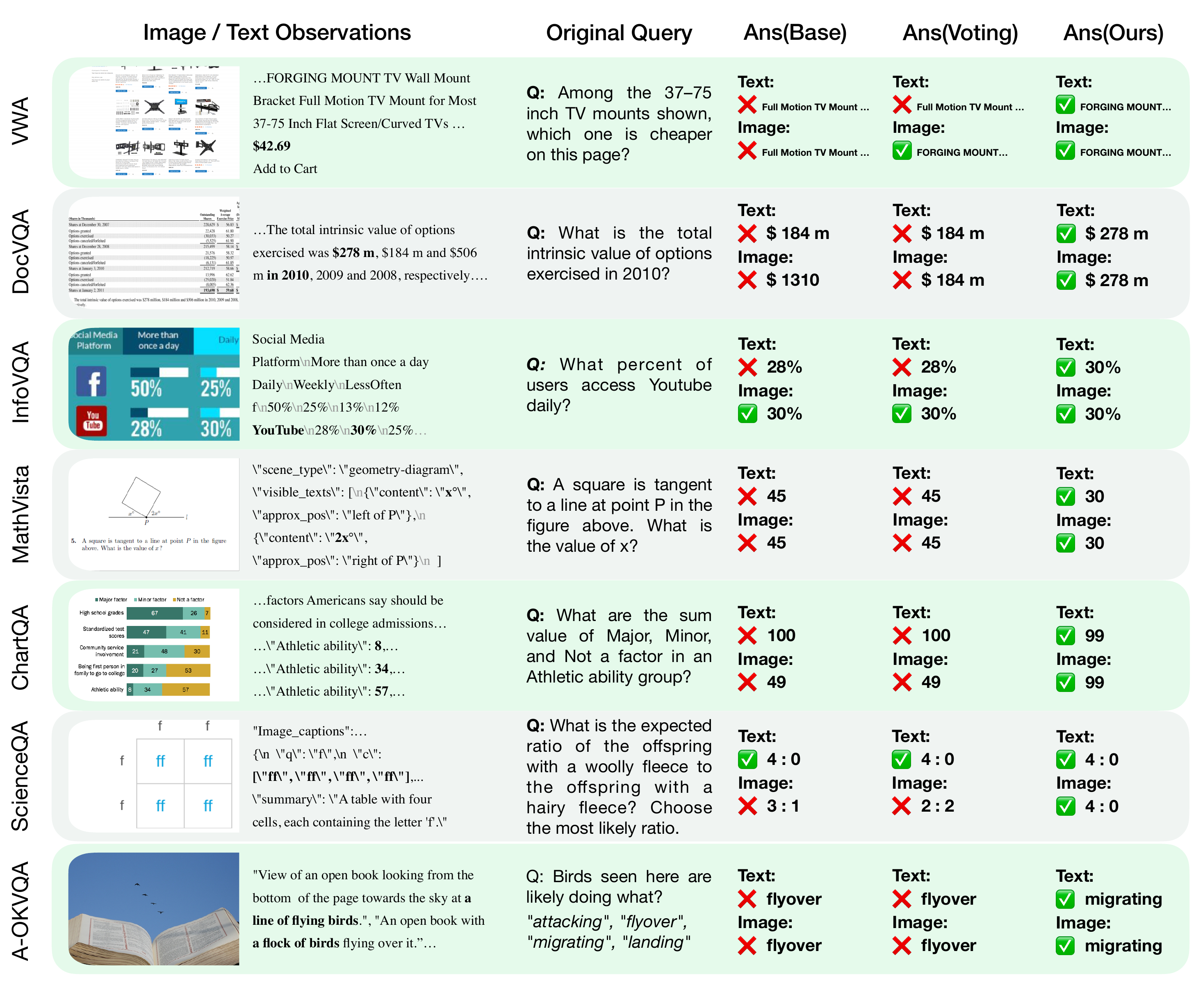}

\caption{
\textbf{Visual comparison among the base model, voting baseline, and \modelname (ours).}
For each example, we display the original multimodal observation, the query, and the predictions from text and image modalities under different methods. 
The base model and voting baseline often yield either conflicting answers across modalities or spurious agreements on incorrect predictions, reflecting the modality gap and unreliable consensus confidence.
In contrast, \modelname enforces cycle-based alignment, producing answers that are both correct and consistent across modalities.
}
\end{figure*}

\subsection{Main Results}
Table~\ref{tab:main_acc} and Table~\ref{tab:main_consistency} summarize the overall performance of \modelname across six multimodal reasoning benchmarks using the Qwen2.5-VL-3B-Instruct and Qwen3-VL-8B-Instruct models. 
We compare our method against the majority vote–based self-improvement method, which for single-modality voting we denote as \textit{Voting (Text-only)}, and for multimodal voting we denote as \textit{Voting (Image+Text)}.

As shown in Table~\ref{tab:main_acc}, \modelname consistently outperforms all voting baselines across datasets and model scales. On ScienceQA, our method improves text and vision accuracy by up to (+7.8/+7.3) points (Qwen2.5-VL-3B) and (+3.0/+1.0) points (Qwen3-VL-8B). Comparable gains appear on ChartQA (+6.1/+2.0) points, MathVista (+6.0/+2.8) points and A-OKVQA (+4.9/+2.7) points, while the complex VWA benchmark sees substantial improvements of (+5.5/+4.2) points. This trend holds on DocVQA and InfoVQA, confirming the cross-modal reward benefits for both modalities.

Our accuracy gain stems from addressing a fundamental flaw in voting baselines. While single modal aggregation improves on the baseline, it fails to model inter-modal relationships. An erroneous rollout can dominate the vote, and even multimodal voting is insufficient in fixing it: it is common that modalities disagree, and the vote simply collapses to the dominant modality. This leaves the underlying conflict unresolved and injects substantial noise into the training signal, limiting accuracy. In contrast, \modelname enforces both finer-grained within-modality and cross-modal consistency. The cycles stabilize unstable reasoning steps and, crucially, resolve inter-modal conflicts, leading to a direct and significant improvement in reasoning capability.

\begin{figure}[t]
  \centering
  \begin{subfigure}[t]{0.48\linewidth}
    \centering
    \includegraphics[width=\linewidth]{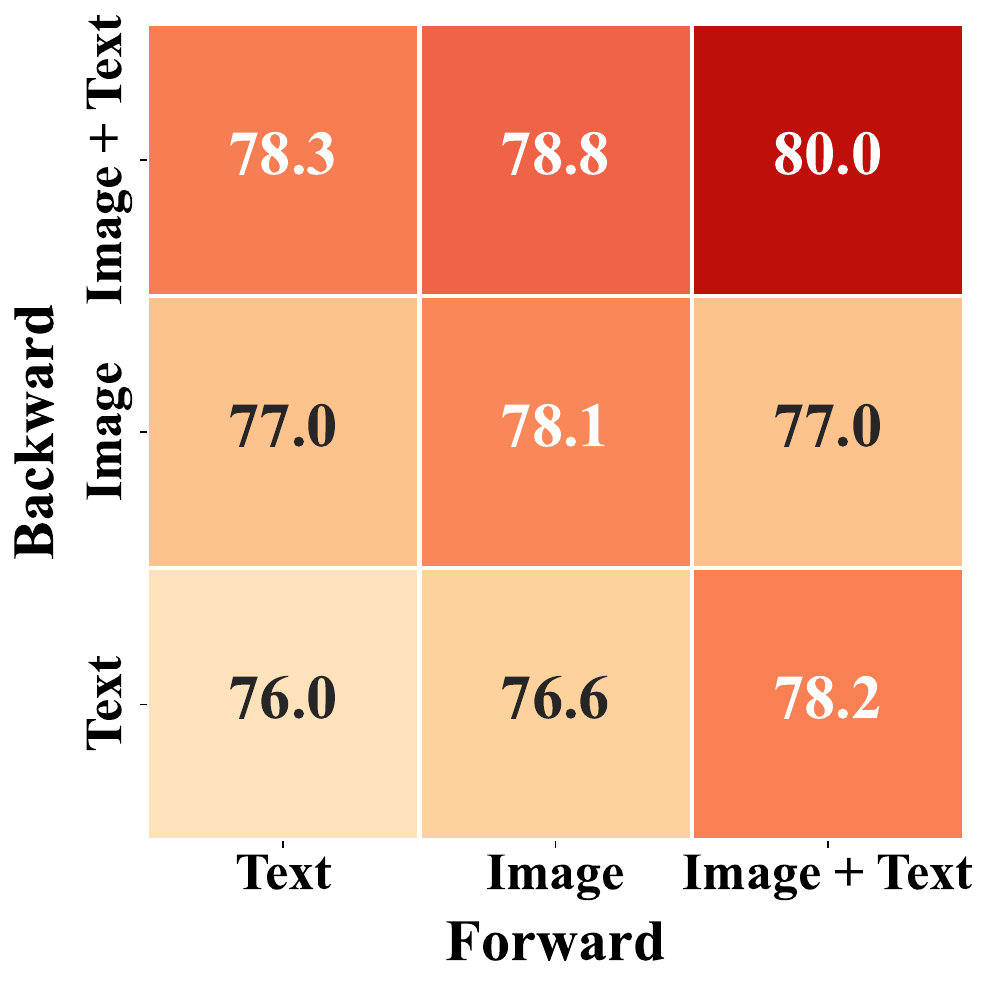}
    \caption{Average accuracy across image and text modalities (\%).}
    \label{fig:sqa_avg_acc}
  \end{subfigure}
  \hfill
  \begin{subfigure}[t]{0.48\linewidth}
    \centering
    \includegraphics[width=\linewidth]{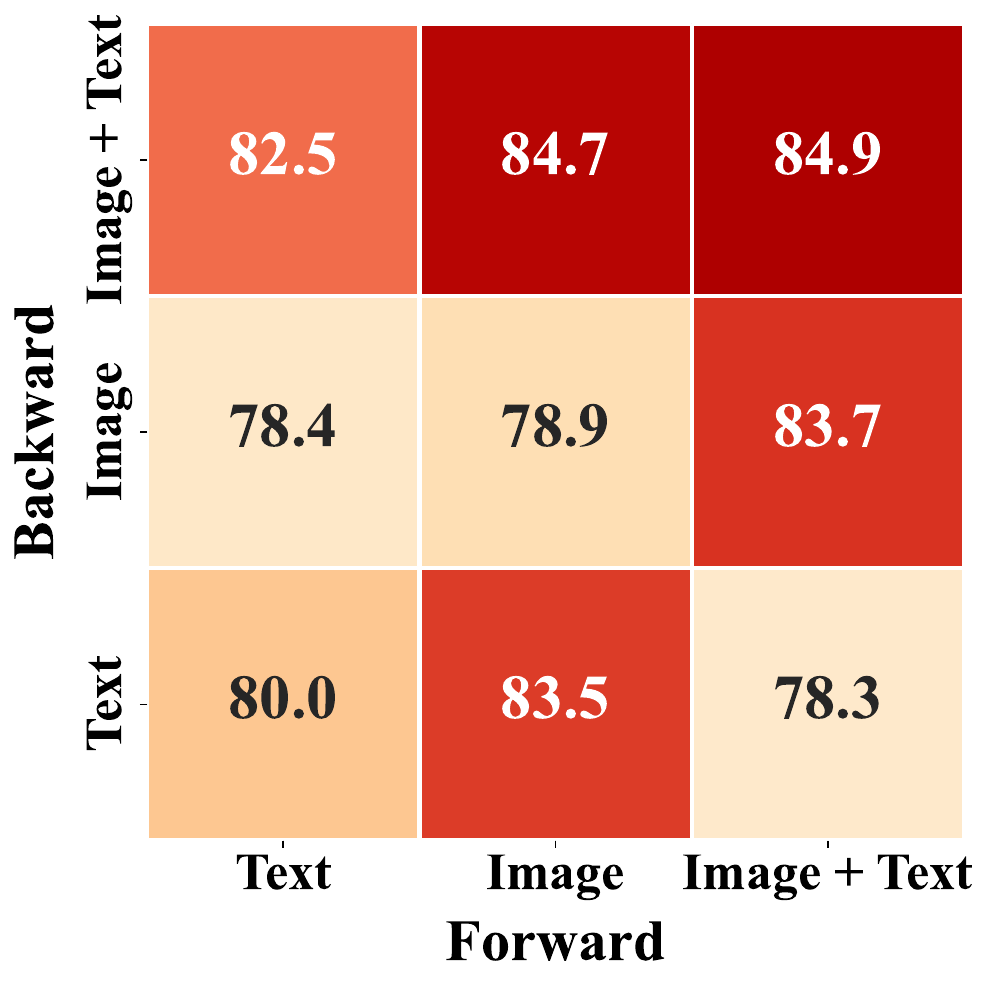}
    \caption{Cross-modal predictions consistency ratio (\%).}
    \label{fig:sqa_consistency}
  \end{subfigure}
  \caption{
  \textbf{Importance of cross modal in cycle consistency path.}
    We show ScienceQA results under \modelname training with nine different forward–backward cycle paths. Backward denotes \textit{answer → query} and forward denotes \textit{query → answer}
    The heatmaps show (a) average accuracy and (b) cross-modal consistency across all modality configurations. 
    Using both image and text modalities for backward query generation consistently leads to higher accuracy and stronger cross-modal agreement, 
    suggesting that multimodal grounding provides a more reliable signal for cycle-consistent reasoning.
    }
  \label{fig:sqa_heatmaps}
\end{figure}

Moreover, Table~\ref{tab:main_consistency} reports the Consistency Ratio—the proportion of samples for which image- and text-based predictions agree. 
\modelname substantially increases agreement across all benchmarks, reaching up to +12.5 points on A-OKVQA, +10.0 points on ScienceQA and +6.1 points on ChartQA with Qwen2.5-VL-3B. 
For the larger Qwen3-VL-8B model, consistency continues to improve (+3–5) points, indicating that \modelname complements model scaling and remains effective even as baseline accuracy increases. 
The gains suggest that enforcing cross-modal cycle consistency not only improves accuracy but also stabilizes multimodal reasoning by aligning both modalities toward a shared semantic interpretation, making the model more robust.

\begin{table*}[t]
\centering
\small
\renewcommand{\arraystretch}{1.05}
\setlength{\tabcolsep}{5pt}
\begin{tabular}{l|ccc|ccc}
\toprule
\multirow{2}{*}{\textbf{Source of Candidate}} 
& \multicolumn{3}{c|}{\textbf{A-OKVQA}} 
& \multicolumn{3}{c}{\textbf{VWA}} \\
\cmidrule(lr){2-4} \cmidrule(lr){5-7}
& Vision Acc & Text Acc & Cons. & Vision Acc & Text Acc & Cons. \\
\midrule
\textbf{Base model}               & 86.1 & 70.3 & 67.8 & 69.0 & 62.9 & 63.9 \\
\textbf{Self-generated answer}    & 87.3 & 71.6 & 77.6 & 74.9 & 65.3 & 68.2 \\
\textbf{Training set answer}      & 88.8 & 75.2 & 80.3 & 74.5 & 67.1 & 67.1 \\
\bottomrule
\end{tabular}
\caption{\textbf{Effect of candidate answer source for initiating the cycle}. 
We compare using the model’s own predicted answer (\textit{Self-generated}) versus using the reference answer from the dataset (\textit{Training-set answer}). 
Results show comparable or better performance without requiring annotated QA pairs, highlighting the self-bootstrapping capability of \modelname.}
\label{tab:candidate_source_results}
\vspace{-3mm}
\end{table*}

\begin{figure}[t]
  \centering
  \includegraphics[width=0.8\linewidth]{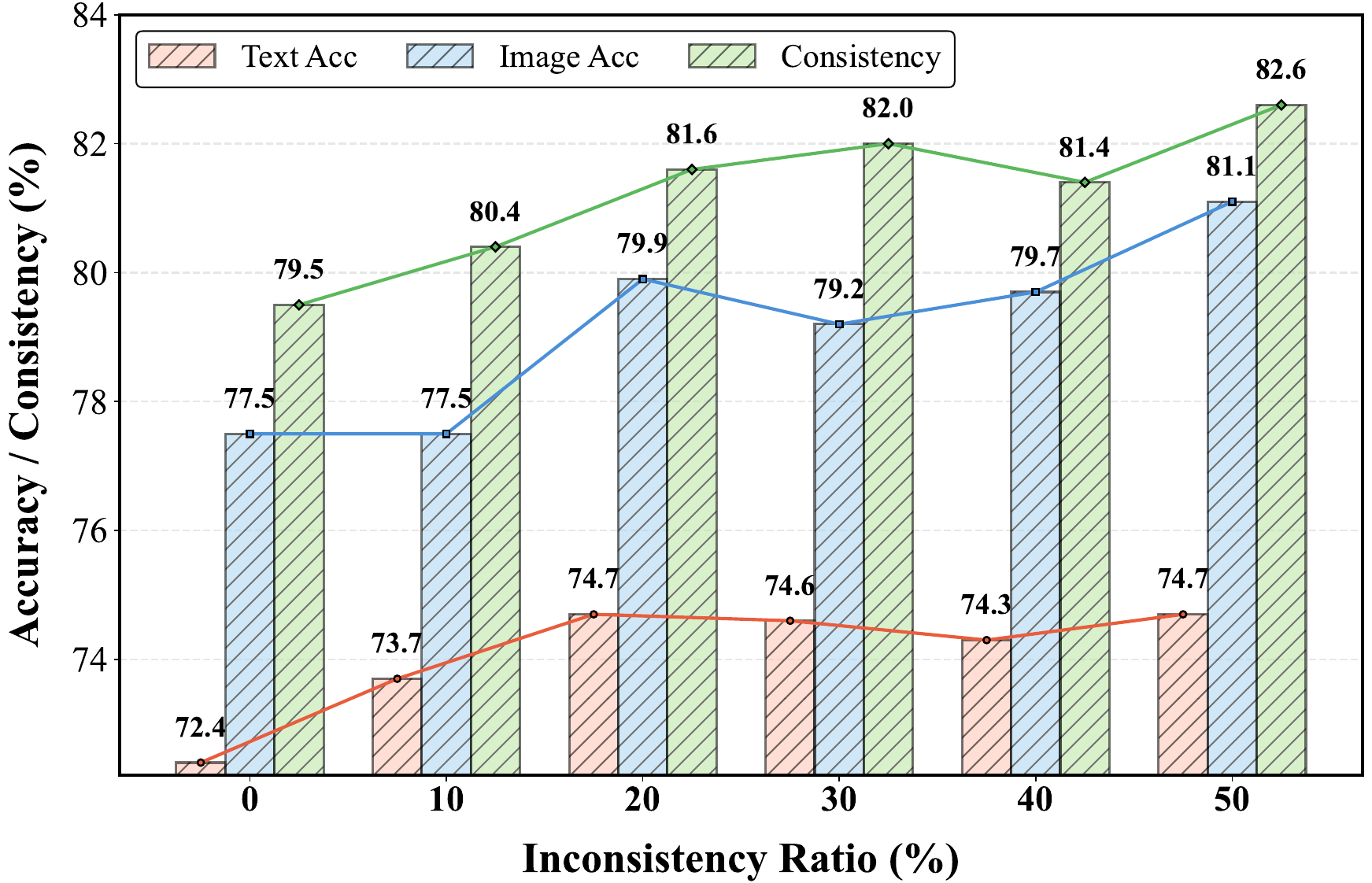}
    \caption{
    \textbf{Effect of cross-modal inconsistency ratio on ScienceQA performance.} 
We fine-tune Qwen2.5-VL-3B-Instruct on ScienceQA with varying fractions of cross-modal inconsistent training samples. 
As the inconsistency ratio increases from 0\% to 50\%, both answer accuracy and cross-modal consistency systematically improve. 
This suggests that exposing the model to more modality conflicts provides a useful structural signal for the RL objective, enabling it to better resolve cross-modal disagreements and yielding more stable multimodal reasoning.
    }
  \label{fig:inconsistency_ratio}
\end{figure}

\subsection{Analysis}
We conduct a series of ablation studies to better understand how each design choice affects the performance of \textbf{\modelname}. 
Our analyses cover cycle path strategies,  how multi-modal inconsistency data affects model reasoning ability, and how to retrieve candidate answers.

\noindent \textbf{Importance of Multimodal Interaction.} We conduct an ablation study to isolate the impact of multimodal interaction. In Table~\ref{tab:cycle_path_interaction_ablation}, we compare three training configurations: 1) single-modality cycle consistency (\textit{Single}), 2) cross-modality cycle consistency (\textit{Cross}), and 3) a combination of both (\textit{Mixed}). Our results show that introducing cross-modal interaction consistently outperforms the single-modality baseline. Moreover, the two objectives prove to be complementary: the \textit{Mixed} variant yields a significant accuracy gain over using either in isolation. Consequently, we adopt the \textit{Mixed} configuration for all subsequent experiments.

\noindent \textbf{Ablation Study of Cycle Paths.} 
We analyze the contribution of specific cycle paths in Figure~\ref{fig:sqa_heatmaps}. 
We report accuracy and consistency across a grid where rows indicate the query-generation modality and columns indicate the answer-prediction modality. 
The ``Image + Text'' entries represent the joint use of both modalities.
We observe that single-path supervision is insufficient; configurations restricted to one path (e.g., only Image$\rightarrow$Image) result in the lowest performance. 
In contrast, incorporating cross-modal paths (e.g., Text$\rightarrow$Image) significantly improves consistency by forcing the model to align semantic representations across modalities. 
Ultimately, the best performance is achieved by the full union of all paths (bottom-right cell). 
This confirms that maximizing the diversity of cycle constraints provides the most robust supervision, effectively mitigating shortcuts and grounding the model's reasoning.

\noindent \textbf{The Importance of Cross-Modal Disagreement Data for Training.} 
We investigate whether training on conflicting data helps or hinders learning. 
We construct controlled ScienceQA subsets (3.2k samples) with fixed size but varying rates of cross-modal inconsistency ($\{0\%, \dots, 50\%\}$). 
Figure~\ref{fig:inconsistency_ratio} shows that as the inconsistency ratio increases, both accuracy and consistency improve. 
This indicates that conflicting samples serve as effective ``hard examples.'' 
By forcing the model to resolve these disagreements via the cycle reward, the RL process achieves tighter cross-modal alignment and more robust reasoning than when training on easy, consistent data.

\noindent \textbf{Source of Candidate Answers.}
We further study how the choice of candidate answers used to start the cycle affects final performance.
We compare two ways of obtaining the candidate answer:
(1) \textit{Self-generated answer}, where the MLLM predicts an initial answer from the input (fully self-supervised),
and (2) \textit{Training-set answer}, where we directly use the reference answer from the dataset (without relying on a labeled query–answer pair).
Both settings are scalable because neither requires human annotation of queries nor majority-vote pseudo labels.

As shown in Table~\ref{tab:candidate_source_results}, the self-generated setting achieves performance nearly identical to the training-set answer setting on web understanding tasks.
This demonstrates that \modelname can reliably bootstrap its own supervision without externally provided answers, highlighting strong potential for scalable, label-free multimodal reinforcement learning.

\section{Conclusion}
Our work shows that robust multimodal reasoning requires models to achieve a rigorous, bidirectional alignment between modalities. 
We start from a key failure in this area---a {modality gap} where MLLMs contradict themselves on visual versus textual views of the same content. 
Instead of merely aggregating conflicting votes, our framework, \modelname, exploits this gap via a cross-modal cycle, which is a powerful untapped resource for self-reward learning. 
\modelname performs backward inference to hypothesize a query, switches modalities, and performs forward inference to reconstruct the answer. By teaching the model to resolve its own internal conflicts, \modelname leads to substantial improvements in both reasoning accuracy and cross-modal alignment. 
Our work suggests that the next generation of robust multimodal reasoning will emerge not just from scaling data, but from enforcing a more rigorous, bidirectional understanding of the world.

\section*{Acknowledgment}
This work used Purdue Anvil GPU through allocation 250774 from the Advanced Cyberinfrastructure Coordination Ecosystem: Services \& Support (ACCESS) program, which is supported by U.S. National Science Foundation grants \#2138259, \#2138286, \#2138307, \#2137603, and \#2138296. We thank Guangxing Han for the insightful discussion. We thank OpenAI for the credit.

\newpage
{
    \small
    \bibliographystyle{ieeenat_fullname}
    \bibliography{main}
}

\clearpage
\setcounter{page}{1}
\maketitlesupplementary

\renewcommand{\thesection}{\Alph{section}}
\setcounter{section}{0}
\renewcommand{\thetable}{A\arabic{table}}
\setcounter{table}{0}

\begin{figure*}[t]
\vspace{-3mm}
  \centering
  \includegraphics[width=0.97\textwidth]{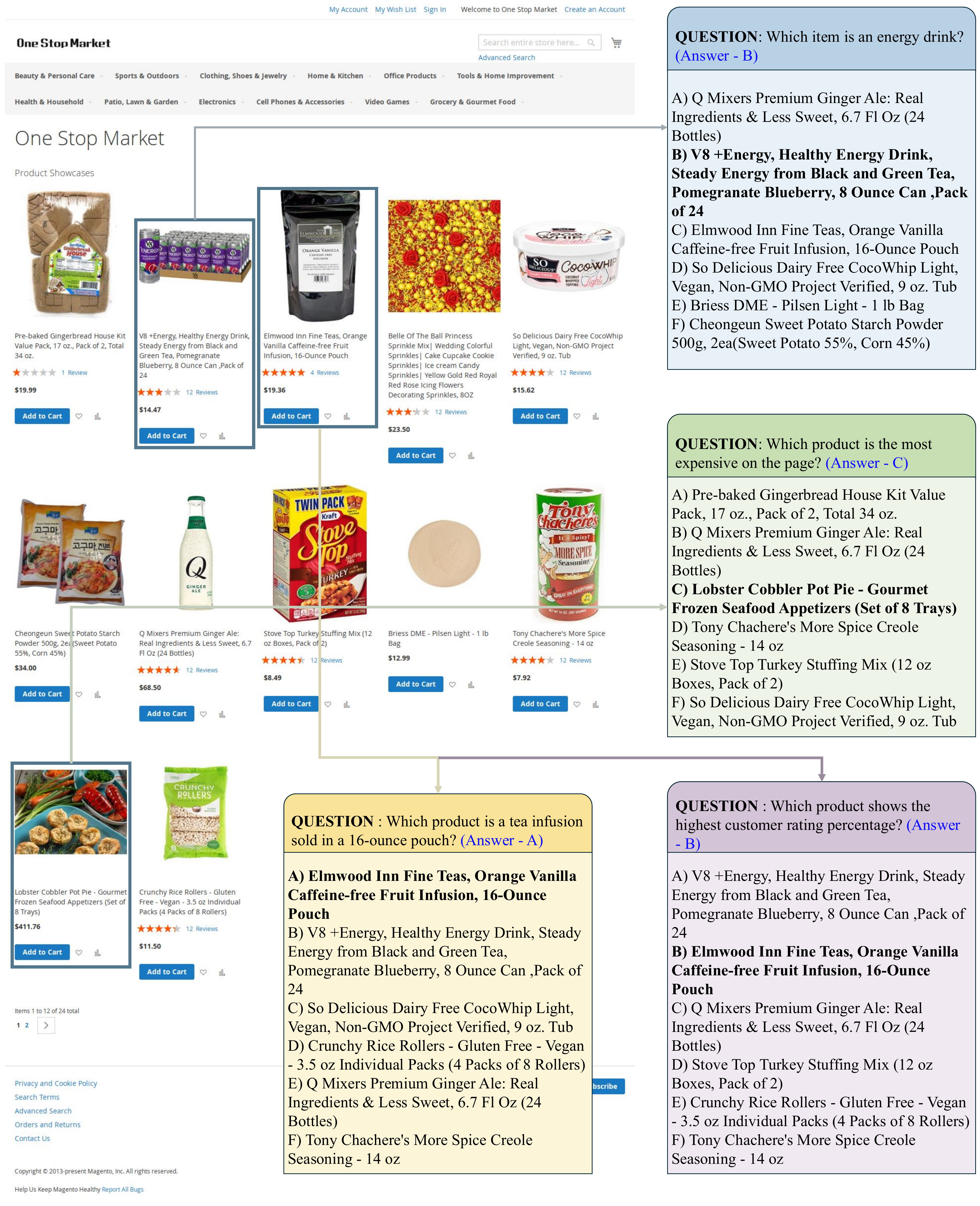}
\caption{
Examples from the reproposed VWA multiple-choice dataset constructed on shopping webpages.
For each webpage, we generate \textbf{10} questions with \textbf{6} answer choices, 
among which \textbf{1} is correct. The questions cover diverse reasoning types, including
product category recognition, price comparison, customer rating analysis, and attribute identification.
}
\label{fig:vwa_1}
\end{figure*}

\begin{figure*}[t]
\vspace{-3mm}
  \centering
  \includegraphics[width=0.97\textwidth]{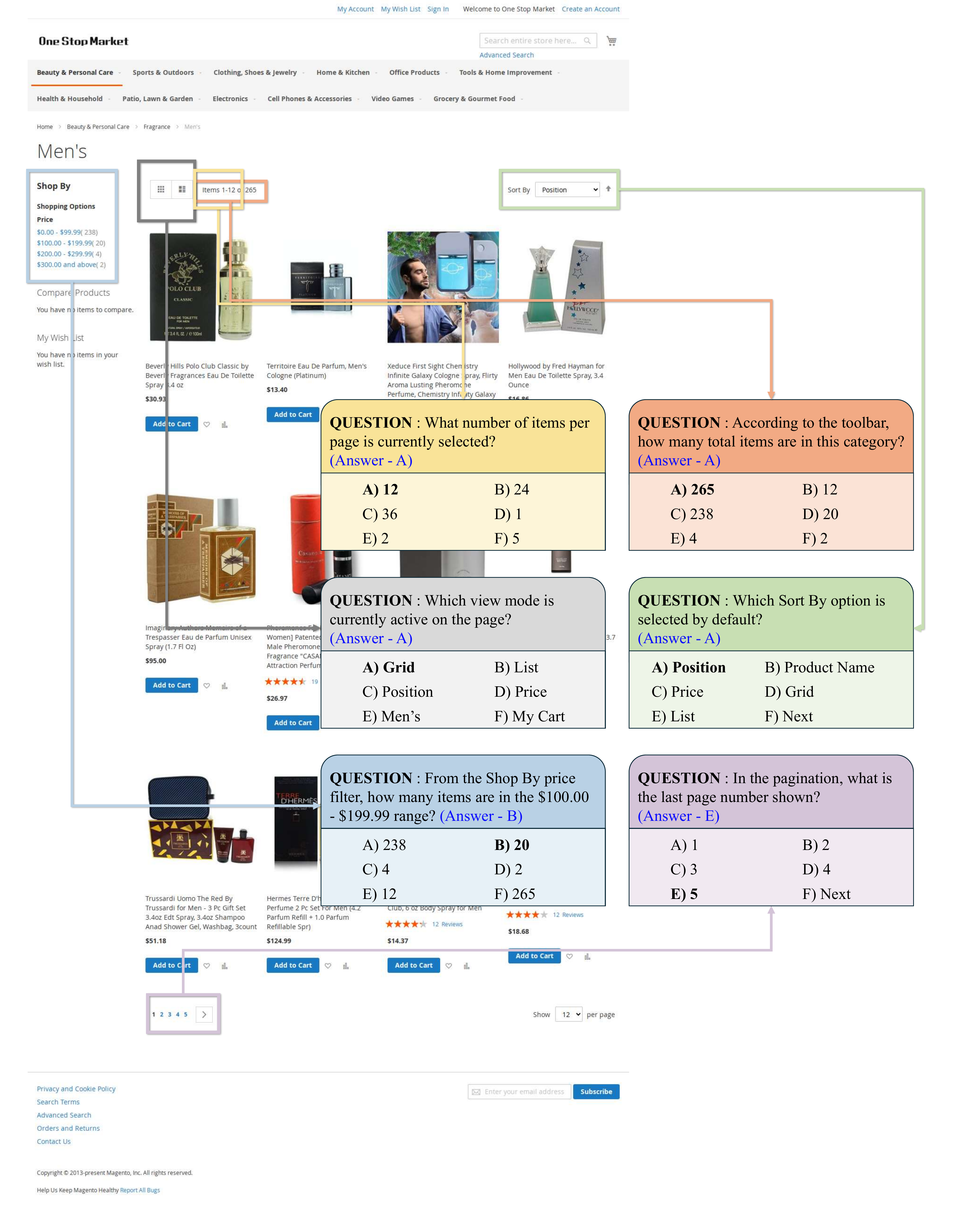}
\caption{
Examples from the reproposed VWA multiple-choice dataset constructed on shopping webpages.}
\label{fig:vwa_2}
\end{figure*}

\section{Training Hyperparameters}
\label{sec:hyperparams}


The key global hyperparameters used throughout all experiments in this paper are listed in Table \ref{tab:train_hyperparams_global}. Unless explicitly stated otherwise, all models were trained under this shared configuration.

\begin{table}[h]
\centering
\begin{tabular}{lc}
\toprule
\textbf{Parameter} & \textbf{Value} \\
\midrule
Global Batch Size & (256, 1024) \\
Learning Rate & $1 \times 10^{-6}$ \\
Weight Decay & $1 \times 10^{-2}$ \\
KL Penalty Coefficient ($\lambda_{\text{KL}}$) & $1 \times 10^{-2}$ \\
Max Steps & 100 \\
Number of Rollouts & 4 \\
Temperature & 1.0 \\
Top-p & 0.95 \\
\bottomrule
\end{tabular}
\caption{Key global hyperparameters used throughout our experiments. Larger batch sizes are adopted when GPU memory permits.}
\label{tab:train_hyperparams_global}
\end{table}

\section{Reproposed VWA Task}

To enable a more straightforward evaluation of multimodal web understanding,
we redesign the VWA dataset into a multiple-choice question answering (MC-QA) task
based on shopping-website pages. For each webpage, we automatically generate
question--answer pairs using GPT-5, followed by human verification to ensure correctness
and quality. Each page contains \textbf{10} questions, and each question provides
\textbf{6} candidate choices, with \textbf{1} correct answer. Visualized examples
are provided in Figures~\ref{fig:vwa_1} and~\ref{fig:vwa_2}.

\section{Prompt Templates}
\label{sec:prompts}

We list the prompt templates used during training and evaluation in Table~\ref{tab:prompt_templates}.

\begin{table*}[h]
\centering
\begin{tabular}{p{0.12\linewidth}|p{0.88\linewidth}}
\toprule
\Large\textbf{Dataset} & \Large\textbf{Prompt Template} \\
\midrule

\multirow{2}{*}{\parbox{1.8cm}{\centering \large\textbf{DocVQA \\ \& \\ InfoVQA}}}
&
{\vspace{1pt}\large\textsc{Backward Query Generation}}\par\medskip

\textbf{System Message:}\par
You are a question generator. Given an observation of a document (infographic) and the correct Answer, produce a single, concise, unambiguous question whose answer is exactly the given Answer and should be asking about the information provided by the observation; when grounded ONLY on the observation provided. Rules: (1) Return ONLY the question text. (2) Avoid ambiguous wording; ensure a 1-to-1 mapping, this means the query should not have answers to it other than the correct Answer provided to you. (3) No extra commentary, quotes, or prefixes. (4) Make sure that the question is asking about the observed document (infographic), don't propose random answers to fit the answer.\par\medskip

\textbf{User Message:}\par
Base rule: You need to generate a question. Here is the structure of the examples:\par\medskip

\textit{few shot examples:}\par\medskip

Example i\par
OBS: \texttt{[TEXT\_OBS]} or \texttt{[IMAGE\_OBS]}\par
Answer: \texttt{[ANS]}\par
\par
Given the observation of the document (infographic), I will come up with a document (infographic) facts based Query that can be answered by the given answer. Query is: \texttt{[QUESTION]}\par
\medskip

\textit{The remaining examples follow the same structure.}\par\medskip

Visual focus: In this task, the OBS is an image version of the document (infographic). You must generate the question based solely on visible information in the image, without assuming or inferring any unseen text or external knowledge.\par
Textual focus: In this task, the OBS is ocr and captioning of an document (infographic) image.\par

OBS: \texttt{[TEXT\_OBS]} or \texttt{[IMAGE\_OBS]}\par
Answer: \texttt{[ANS]}\par
Given the observation of the document (infographic), I will come up with a document (infographic) facts based Query that can be answered by the given answer. Query is:\par\medskip

\textbf{Assistant Message:} (Generated response)\medskip
\\ \cline{2-2}

&
{\vspace{1pt}\large\textsc{Cycle Verification}}\par\medskip

\textbf{User Message:}\par
You are a helpful assistant for document VQA. Answer with the exact final answer only.\par
\par
OBS: \texttt{[TEXT\_OBS]} or \texttt{[IMAGE\_OBS]}\par
Query: \texttt{[QUESTION]}\par
Answer concisely with only the final answer.\par\medskip
\textbf{Assistant Message:} (Generated response)\medskip
\\
\bottomrule
\end{tabular}
\caption{\textbf{Prompt templates for each dataset and prompt type.} 
The \textsc{Backward Query Generation} prompt infers plausible multimodal queries conditioned on each candidate answer; 
and the \textsc{Cycle Verification} prompt assesses whether forward inference from these queries reconstructs the original answer, providing the cycle-consistency supervision signal.}
\label{tab:prompt_templates}
\end{table*}

\begin{table*}[h]
\centering
\begin{tabular}{p{0.12\linewidth}|p{0.88\linewidth}}
\toprule
\Large\textbf{Dataset} & \Large\textbf{Prompt Template} \\
\midrule

\multirow{2}{*}{\parbox{1.8cm}{\centering \large\textbf{ChartQA}}}
&
{\vspace{1pt}\large\textsc{Backward Query Generation}}\par\medskip

\textbf{Image-specific part:}\par
\texttt{[IMAGE\_OBS]} \par
You are writing ONE short, clear, self-contained question about a chart image such that the answer equals a GIVEN TARGET ANSWER.\par\medskip

\textbf{Text-specific part:}\par
You are writing ONE short, clear, self-contained question about a chart based ONLY on the following caption text. The question must have the GIVEN TARGET ANSWER. \texttt{[TEXT\_OBS]}: \par\medskip

\textbf{Shared rules and ending:}
Rules: \par
- Must be answerable using chart visual information only. \par
- Must not reveal or restate the answer. \par
- Avoid yes/no, multiple choice, multi-part questions. \par
- Include necessary qualifiers (series, category, unit, timestamp). \par
- Length: 1--2 sentences, no explanation. \par
Target Answer: \texttt{[ANSWER]}. Write the question now. Return only the question string.

\medskip
\\ \cline{2-2}

&
{\vspace{1pt}\large\textsc{Cycle Verification}}\par\medskip

\textbf{Image-based QA Prompt:} \par
\texttt{[IMAGE\_OBS]} \par
You are given a chart image and a question. Use ONLY the information that is explicitly visible in the chart (titles, labels, legends, tick marks, data labels, notes). \par

Question: \par
\texttt{[QUESTION]} \par

Answer concisely with plain or numeric text only (no reasoning, no steps, no formatting).
\medskip

\textbf{Text-based QA Prompt:}\par
You are given a chart description in JSON extracted from an image. Answer the user's question using ONLY the information contained in the JSON. \par

JSON: \par
\texttt{[TEXT\_OBS]} \par

Question: \par
\texttt{[QUESTION]} \par

Answer concisely with plain or numeric text only (no reasoning, no steps, no formatting).
\par\medskip
\\
\bottomrule
\end{tabular}
\caption{\textbf{Prompt templates for each dataset and prompt type.} 
The \textsc{Backward Query Generation} prompt infers plausible multimodal queries conditioned on each candidate answer; 
and the \textsc{Cycle Verification} prompt assesses whether forward inference from these queries reconstructs the original answer, providing the cycle-consistency supervision signal.}
\label{tab:prompt_templates}
\end{table*}

\end{document}